\documentclass[11pt,a4paper]{article}

\usepackage[utf8]{inputenc}
\usepackage[T1]{fontenc}
\usepackage{lmodern}
\usepackage[margin=1in]{geometry}

\usepackage{graphicx}
\usepackage[table]{xcolor}
\usepackage{booktabs}
\usepackage{tabularx}
\usepackage{subcaption}
\usepackage{amsmath,amssymb}
\usepackage{xspace}
\usepackage{tipa}
\usepackage{microtype}
\usepackage[onehalfspacing]{setspace}
\usepackage[round]{natbib}
\usepackage{url}
\usepackage{hyperref}

\let\cite\citep

\graphicspath{{figs/}{./}}

\newcommand{\PreAF}{Pre-AF~13\xspace}
\newcommand{\PreAFnine}{Pre-AF~9\xspace}

\title{\bfseries \PreAF: An Interpretable Atrial Fibrillation\\ Risk Score Mined from Discharge Reports}

\author{\begin{minipage}{0.95\textwidth}\centering
Olga Shakhmatova\,$^{1,\dagger,*}$,
Dmitrii Kriukov\,$^{2,3,\dagger}$,
Daniil Larionov\,$^{4}$,
Nikita Khromov\,$^{5}$,
Iaroslav Bespalov\,$^{3}$,
Alexander Zolotarev\,$^{6}$,
Kirill Grishchenkov\,$^{7}$,
Ekaterina Ivanova\,$^{2}$,
Miron Kuznetsov\,$^{2}$,
Ilya Sochenkov\,$^{7,8,9,10}$,
Elizaveta Panchenko\,$^{1,\ddagger}$,\\
Artem Shelmanov\,$^{11,\ddagger}$,
and Dmitry V. Dylov\,$^{2,3,\ddagger,*}$
\\[1.5ex]
\scriptsize
$^{1}$National Medical Research Center of Cardiology named after Academician E.I. Chazov, Moscow, Russia\\
$^{2}$Skolkovo Institute of Science and Technology (Skoltech), Moscow, Russia\\
$^{3}$Artificial Intelligence Research Institute (AIRI), Moscow, Russia\\
$^{4}$University of Mannheim, Mannheim, Germany\\
$^{5}$Russian Center for Scientific Information (RCSI), Moscow, Russia\\
$^{6}$Institute of Cyber Intelligence Systems, National Research Nuclear University MEPhI, Moscow, Russia\\
$^{7}$M.V. Lomonosov Moscow State University, Moscow, Russia\\
$^{8}$Institute for Information Transmission Problems of the Russian Academy of Sciences (Kharkevich Institute), Moscow, Russia\\
$^{9}$Ivannikov Institute for System Programming of the Russian Academy of Sciences (ISP RAS), Moscow, Russia\\
$^{10}$Federal Research Center ``Computer Science and Control'' of the Russian Academy of Sciences (FRC CSC RAS), Moscow, Russia\\
$^{11}$Mohamed bin Zayed University of Artificial Intelligence (MBZUAI), Abu Dhabi, United Arab Emirates\\[0.5ex]
$^{\dagger}$Co-first author \quad
$^{\ddagger}$Co-senior author \quad
$^{*}$Corresponding authors: olga.shahmatova@gmail.com, d.dylov@skol.tech
\end{minipage}
}

\date{}

\begin{document}

\maketitle

\begin{abstract}
\noindent\textbf{Background.} Atrial fibrillation (AF) is the most prevalent cardiac arrhythmia and a major determinant of patient prognosis. Established AF risk scores rely on common factors such as older age and hypertension, which are nearly ubiquitous among patients with existing cardiovascular disease (CVD) and therefore offer limited stratification in this high-risk group. Most existing tools also target long-term (5-10 year) rather than medium-term prediction. We developed interpretable machine-learning models predicting AF risk over both a 24-month and an entire follow-up horizon in CVD patients, using routinely collected hospital data.

\medskip
\noindent\textbf{Methods.} In this single-center retrospective study, we analyzed electronic health records (EHRs) from the National Research Cardiology Center (Russia) for patients aged $\geq$18 years with documented CVD but without pre-existing AF, who were hospitalized more than once between January 2012 and May 2019. A custom natural language processing pipeline transformed unstructured discharge reports into 73 structured features, combining a rule-based parser with transformer-based named-entity recognition, validated against manually annotated test sets. Using the LightAutoML framework, we built a full model (all 73 features), a simple model (reduced subset), and a linear model for bedside risk scores. Performance was assessed by ROC AUC, compared with CHARGE-AF, C2HEST, MHS, and HAVOC scores, and interpreted using SHAP analysis.

\medskip
\noindent\textbf{Results.} Of 80,576 records from 45,000 unique patients, 17,562 met inclusion criteria; 1,438 (8.19\%) developed AF (median age 62.0 years; 36.4\% women). The NLP pipeline achieved high performance on the tasks of feature extraction. On the test set, the full model reached a ROC AUC of 0.735 (24 months follow-up target) and 0.696 (entire follow-up target), with the simple model nearly identical (0.725 and 0.696); all non-linear models outperformed the four established clinical risk scores (ROC AUC 0.53-0.64). The simple model comprises 13 features and is therefore named \PreAF. SHAP identified age and left atrial volume as dominant predictors. A conventional risk score (\PreAFnine) based on a linear model was also developed, which stratified the observed 24-month incidence of AF from approximately 7\% to 36\%.

\medskip
\noindent\textbf{Conclusion.} Interpretable machine-learning models built from routinely collected EHR data can identify high-AF-risk patients with CVD, outperforming clinical risk scores.

\end{abstract}

\section{Introduction}

Atrial Fibrillation (AF), the most common cardiac arrhythmia \cite{martin2024}, is a significant contributor to the risk of stroke, heart failure, chronic kidney disease, dementia, and mortality \cite{odutayo2016atrial, zuin2021risk}. The substantial burden of complications associated with AF is partly attributable to delays in timely intervention, especially among patients exhibiting nonspecific symptoms or remaining entirely asymptomatic \cite{tayal2008atrial}. Identifying individuals at high risk of AF could enable the targeting of systematic screening efforts, ensuring they are both clinically effective and economically viable \cite{hlatky2009criteria}.

Proactive management of comorbidities and the implementation of upstream therapy may mitigate the risk of AF development \cite{emdin2015effect, hoskuldsdottir2021potential}. A personalized approach to AF risk assessment, focusing on patient-specific risk factors, holds promise for enhancing the efficacy of primary prevention strategies, including improving patient adherence to treatment and lifestyle modifications.

To assess AF risk, numerous risk scores have been developed that incorporate clinical, instrumental, and laboratory parameters \cite{alonso2013simple, li2019simple, chamberlain2011clinical, aronson2018risk, hamada2019simple, kokubo2017development, zolotarev2020optical}. However, these tools have various limitations. For example, clinical risk scores designed for the general population, which rely on common risk factors such as age, hypertension, smoking, and obesity, may be less effective in patients with established cardiovascular disease (CVD). In this population, these risk factors are nearly ubiquitous, limiting their utility for further risk stratification. A more refined approach is needed for patients with known CVD, one that incorporates markers of structural or electrical remodeling of the left atrium, as well as clinical scenarios that predispose such remodeling. Furthermore, existing risk stratification methods often overlook the impact of ongoing treatments. For instance, while polyunsaturated fatty acids are associated with an increased risk of AF \cite{gencer2021effect}, SGLT2 inhibitors \cite{sfairopoulos2023association, zheng2022association} and mineralocorticoid receptor antagonists \cite{neefs2017aldosterone,pabon2025finerenone} appear to reduce this risk. 

Additionally, past invasive interventions, which may contribute to the development of an arrhythmogenic substrate in the left atrium, are rarely involved in risk assessments. It is also worth noting that current risk scores are primarily designed for medium-term AF prediction (5-10 years) \cite{alonso2013simple, li2019simple, aronson2018risk, hamada2019simple, kokubo2017development}. While this time frame is appropriate for identifying targets for preventive interventions, it is less suitable for identifying patients who require immediate screening. The factors associated with a high risk of AF in the short term (6–24 months) may differ from those relevant over longer periods, necessitating tailored predictive approaches based on specific clinical objectives.

Advances in understanding the genetic predisposition to AF have led to the development of polygenic and clinical-polygenic risk scores \cite{okubo2020predicting, marston2023polygenic, muse2018validation}. However, their application in clinical practice remains limited by the accessibility of genetic testing. 

Contemporary AI-based predictive models \cite{pipilas2023use, shvetsov2020unsupervised}, while powerful, often lack interpretability for both clinicians and patients, diminishing their acceptance compared to traditional risk scores based on conventional statistical methods.
Our study aimed to develop AI-based models capable of predicting AF risk over both short-term (24 months) and longer-term horizons for patients with established cardiovascular disease, while maintaining interpretability and identifying the most significant predictors.

\section{Materials and Methods}

\subsection{Data Sources and Study Population}
This retrospective observational study utilized data from the hospital information system of the National Research Cardiology Center (Russia), which serves multiple regions across the country. The study population included patients aged 18 years and older with documented cardiovascular pathology but without pre-existing atrial fibrillation (AF) who were hospitalized more than once between January 2012 and May 2019.

\subsection{Data Collection and Primary Outcomes}

Using Natural Language Processing (NLP) techniques, we transformed a large corpus of unstructured electronic health records (EHRs) into a structured set of well-defined features (see the supplementary materials for the quality assessment of features). Data extraction was performed from the following sections of discharge summaries: diagnosis, physical examination, laboratory tests, instrumental tests, and medication recommendations. We evaluated established risk factors for AF, including those incorporated in existing risk scores, markers of atrial cardiomyopathy, and factors that may contribute to the development of an arrhythmogenic atrial substrate or act as AF triggers, such as cardiac interventions and pharmacological treatments.

\subsection{Natural Language Processing Pipeline}

We extract features from EHRs in two ways:
(1) using a rule-based parser and
(2) using a deep learning–based information extraction model. 
The first group encompasses features that are relatively easy to extract with straightforward methods such as context-free grammars or even regular expressions. This group, for example, includes features such as \emph{age}, \emph{gender}, \emph{height / weight / body mass index}, \emph{presence of diabetes}, among others. 
The second group includes features that are difficult to extract with a rule-based approach due to the complexity and variability of natural language in EHRs. Machine learning helps to address this problem by mining deep patterns from annotated datasets. This group of features includes: \emph{PAD}: peripheral artery disease, \emph{Medication}: medication consumed by the patient, and \emph{LVH}: left ventricular hypertrophy.

\subsubsection{Rule-Based Parser}

The rule-based parser is based on regular expressions and a GLR parser that is able to efficiently process texts generated by context-free grammars: Yargi Parser \cite{yargy-parser}.
For constructing rules for feature extraction, we first asked medical experts to list descriptions of all possible mentions of the features. We manually annotated a small dataset of approximately 1,000 fully labeled texts and used it for the development of rules.

For each feature, we manually developed a dedicated extraction rule, which was then evaluated on a validation dataset.
If the performance score was low, medical experts reviewed the errors, updated the documentation as needed, and revised the rule accordingly. This process was repeated iteratively until satisfactory performance was achieved.
If a rule achieved a high score, medical experts reviewed a random sample of EHR documents from the full dataset to validate the correctness of the rule-based approach.
For simple features -- for example, extracting the presence of hypertension, rules required minimal revision. However, for more complex features, the process often required 8 to 15 iterations of revisions.

\subsubsection{Information Extraction Based on Deep Learning and Active Learning}

Developing text processing applications with machine learning typically requires large volumes of labeled data. Although many annotated corpora exist for resource-rich languages, they rarely address the specific demands of real-world use cases. In cardiology, extracting information from EHRs involves domain-specific clinical concepts and documentation practices that are poorly represented in public datasets. As a result, manual annotation becomes necessary -- a process that is both time-consuming and resource-intensive. Crowdsourcing is not a feasible solution in this context due to the sensitive nature of EHRs and the need for annotations to be performed by highly trained physicians. This further increases the cost and complexity, especially when annotation guidelines are intricate and require deep clinical expertise.

To address the complexity of annotation in this project, we use active learning (AL) \cite{settles2009active}. This technique prioritizes the most informative examples for annotation based on the uncertainty of an intermediate model trained on the instances labeled so far. The typical AL pipeline proceeds iteratively: we re-train the intermediate model, apply it to the unlabeled pool, and select the most uncertain predictions for human annotation. Once experts provide the labels, the next iteration begins with the expanded training set. By focusing expert effort where it is most needed, AL achieves strong information extraction performance with far less labeled data compared to exhaustive annotation, saving the limited and costly time of medical experts \cite{shelmanov2021active,tsvigun-etal-2022-towards,tsvigun-etal-2022-altoolbox}.

\subsubsection{Data Annotation for Testing}

To evaluate the performance of both types of features, those extracted by the rule-based parser and those extracted by the machine-learning model, we annotated dedicated test sets. Specifically, we selected 100 electronic health records and manually annotated their discharge summaries. Active learning was not used during test-set annotation in order to mitigate potential bias. The dataset was annotated using the BRAT annotation tool \cite{stenetorp-etal-2012-brat}.

\subsection{Target Variables}

The task is formulated as a prediction of whether AF occurs within a predefined risk interval (not as time-to-event prediction). 
The target variable (endpoint) was constructed in two ways: the 24-month follow-up target and the entire follow-up period target.

\textbf{24-month follow-up target.} The target is a binary indicator of whether AF is first diagnosed within a fixed risk window relative to the index visit. Prediction is assumed to be made at the time of a given visit, and the window spans from one week to 24 months after that visit (\textit{i.e.}, the interval is [visit + 1 week, visit + 24 months]). The one-week lower bound excludes AF documented essentially concurrently with the index visit, which would represent prevalent rather than incident disease.

For each index visit, we searched the subsequent patient records for an AF diagnosis within the corresponding risk window. If AF was diagnosed at any follow-up visit within this window, the index visit was assigned a target value of 1. If no AF was diagnosed at any visit within the window, the index visit was assigned a label of 0. A label of 0 was also assigned when there were no visits within the window itself, provided that a later visit occurred and the first such visit recorded no AF diagnosis.

If the available follow-up records did not allow us to determine whether AF occurred within the risk window, the instance was excluded from the training and test data. This applied to two situations: cases with no subsequent visits and cases in which no visit occurred within the risk window, but the first visit after the window recorded an AF diagnosis. The latter case is excluded because the exact onset of AF was unknown and may have occurred outside the observed interval.

\begin{figure}[htbp]
    \centering
    \includegraphics[width=\linewidth]{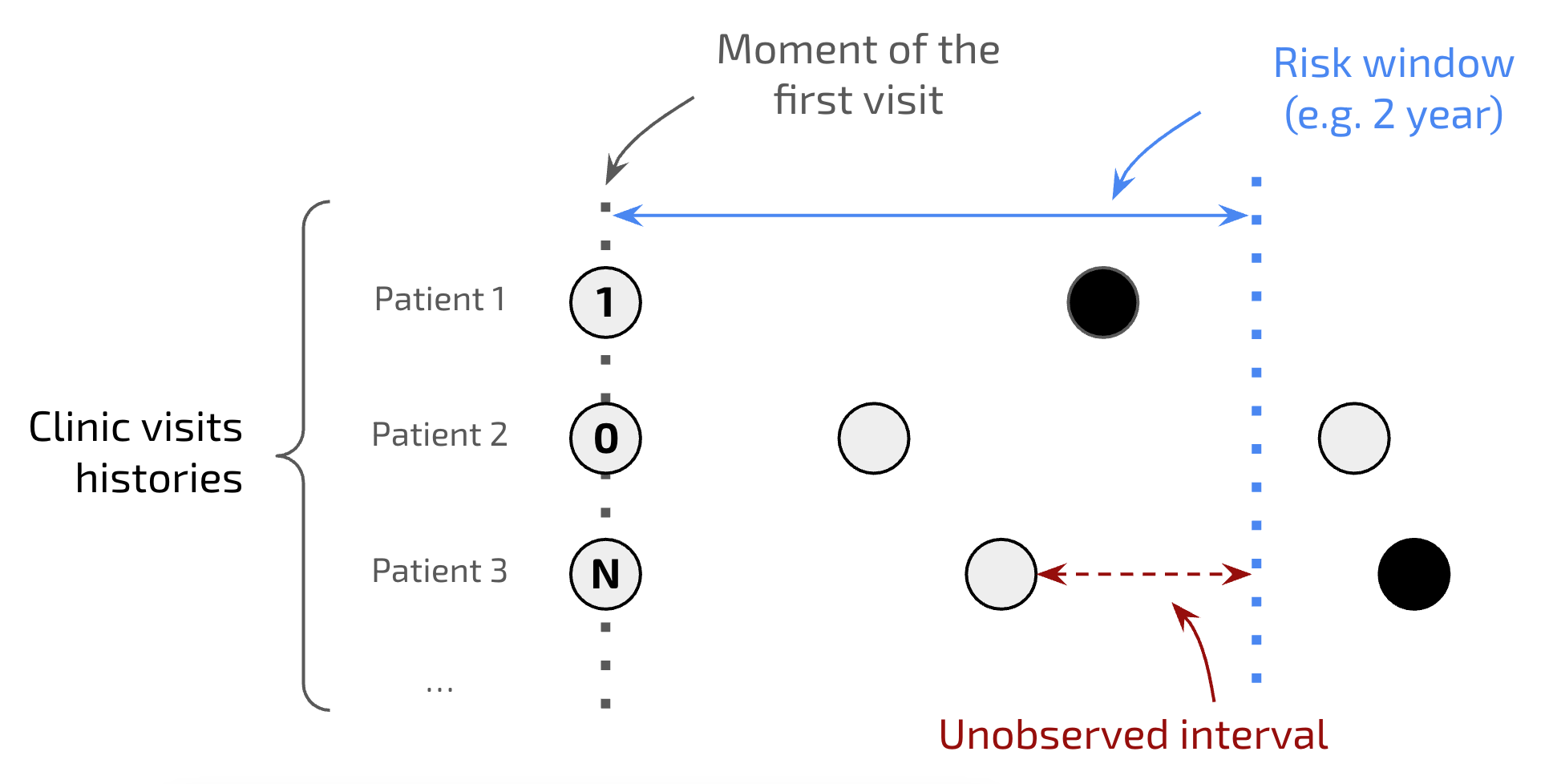}
    \caption{The illustration of target variable construction principle. Black circles show the cases when AF was diagnosed during the visit, while gray circles demonstrate the absence of diagnosis during a visit. The label (1, 0 or N) is assigned based on the follow-up history of visits. Importantly, we assign the label to be N (i.e. NaN) in case the diagnosis has been first made outside the risk window.
    }
    \label{fig:target_construction}
\end{figure}

See Figure~\ref{fig:target_construction} for an illustration of the target construction procedure.

\textbf{The entire follow-up period target.} This target is a binary indicator of whether AF was first diagnosed at any visit after the index visit. A visit was labeled 1 if AF was subsequently diagnosed at any point during follow-up, and 0 if no AF diagnosis was recorded at any follow-up visit. If the patient had no follow-up visits after the index visit, the outcome could not be determined, and the visit is excluded from the training and test data. To avoid counting AF identified essentially concurrently with the index visit, just as with the 24 month target, we do not take into account subsequent visits within a one week period from the index visit.

\subsection{Feature Vector Construction}

\subsubsection{Feature Preprocessing}

\begin{figure}[htbp]
    \centering
    \includegraphics[width=1.0\linewidth]{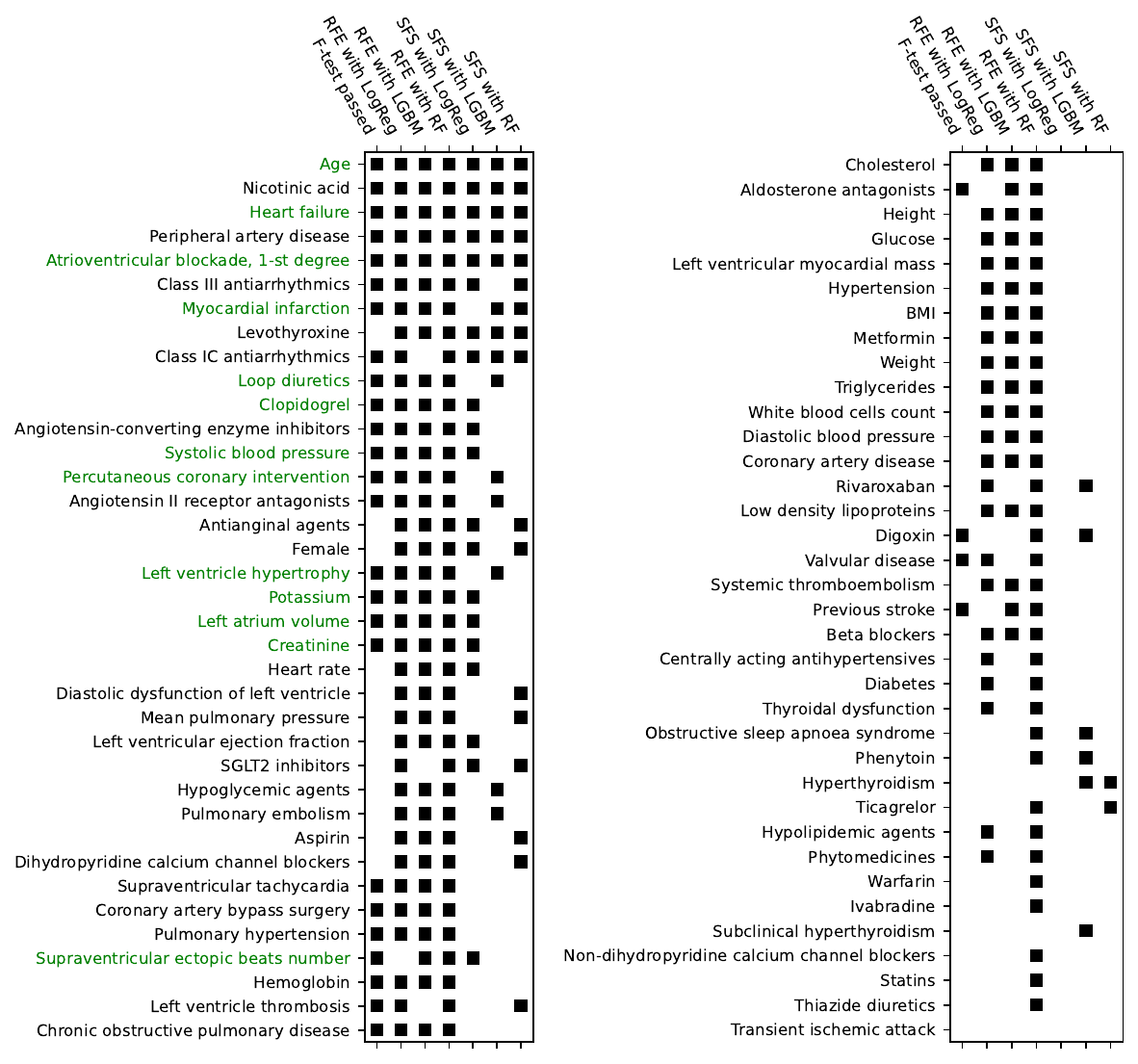}
    \caption{Results of the feature selection procedures used for constructing the ``simple'' model. The right panel is a vertical continuation of the left panel. Each column corresponds to one of the seven feature selection tests: the F-test (univariate), Recursive Feature Elimination (RFE) with logistic regression, LightGBM (LGBM), and random forest (RF), and Sequential Feature Selection (SFS) with the same three classifiers. Each column represents a feature. Black squares indicate that the feature was selected by the corresponding method. A feature was included in the final shortlist if it passed the F-test and was selected by at least one model in both RFE and SFS procedures.}
    \label{fig:feature_selection}
\end{figure}

In total, 73 features were extracted after the processing of EHRs (see Figure \ref{fig:feature_selection} for the full list of features). The dataset included numerical (e.g., BMI, age) and binary features (e.g., clopidogrel administration, previous myocardial infarction). Outliers in numerical features were filtered out. Missing values in binary features corresponding to drug administrations were filled with NaNs. Missing values for linear model construction (see below) were imputed using the k-Nearest Neighbors method. Missing values for models constructed using AutoML 
were not imputed due to the internal ability of AutoML to handle these values automatically through a mixture of different approaches. For the linear model, features also underwent the Min-Max scaling procedure. Negative mentions (e.g., ``ruled out stroke'') were excluded from the analysis.

\subsubsection{Statistical Analysis of Features}

Patient characteristics were reported as median (interquartile range, IQR) for continuous variables and as counts (percentages) for categorical variables. Continuous variables were compared using the Mann-Whitney U test, while categorical variables were compared using Fisher’s exact test.

\subsection{Model Construction}

Our work pursues two goals: (1) to develop a high-performing predictive model (full model) that establishes an upper bound on achievable performance for diagnosing AF, and (2) to develop a practical model (simple model) that can be readily used by physicians in clinical practice. 

The dataset was partitioned at the patient level: 84\% of unique patients were allocated to the derivation (training) set and the remaining 16\% to the validation (test) set. Because the split was performed on patient identifiers rather than individual visits, all records belonging to a given patient were assigned to the same subset, ensuring no patient overlap between training and test data and preventing information leakage across splits. The training set was class-balanced by random under-sampling of the majority (AF-negative) class, whereas the test set retained the natural class distribution to provide an unbiased estimate of performance.

\subsubsection{Construction of the Full Model}

The full AF diagnostic model was trained using all 73 features extracted from EHRs (Figure \ref{fig:feature_selection}). We employed the LightAutoML framework \cite{vakhrushev2021lightautoml}, which implements a complete machine learning pipeline, including data preprocessing (e.g., imputation of missing values and feature scaling), automated model selection, hyperparameter tuning, and cross-validation. This pipeline combines and stacks multiple algorithms to improve performance and generalizability. Specifically, the ensemble included random forest, LightGBM (LGBM), L2-regularized linear regression, and CatBoost \cite{dorogush1810catboost}, using a combination of stacking and bagging strategies. The area under the receiver operating characteristic curve (ROC AUC) was used as the primary optimization metric. Two separate models were trained to predict outcomes over different time horizons: one for a 24-month follow-up period and another for the entire available follow-up period.

\subsubsection{Construction of the Simple Model}

The ``simple'' model is designed for practical use by medical professionals, particularly in settings where manually entering a large number of input features is impossible due to time constraints or limited data availability. We construct a compact model using a smaller, interpretable subset of features selected through a multi-step process that combines statistical and model-based approaches.

We applied three feature selection strategies: a univariate F-test (based on the ``f\_classif'' function from scikit-learn), Sequential Feature Selection (SFS), and Recursive Feature Elimination (RFE). While the F-test evaluates the individual statistical association between each feature and the target variable, SFS and RFE are model-driven methods that assess feature importance in the context of a predictive model. To improve robustness and account for model variability, both SFS and RFE were applied using logistic regression, LightGBM (LGBM) \cite{ke2017lightgbm}, and a random forest as baseline models.

In total, seven feature selection procedures were performed: one using the F-test and three each using SFS and RFE with different classifiers. A feature was included in the final shortlist for the ``simple'' model only if it met all of the following criteria: it demonstrated statistical significance in the F-test, was selected by at least one model in the SFS procedure, and was selected by at least one model in the RFE procedure. This consensus-based approach ensured that the selected features were consistently relevant across different selection techniques and model types, resulting in a minimal yet informative feature set suitable for clinical application. As a result, 21 of 73 features satisfied the defined criteria (Figure \ref{fig:feature_selection}).
However, after additional manual inspection, only 13 of the 21 features were selected for the ``simple'' model construction.

The same LightAutoML pipeline was applied to train a model on the shortlisted set of features using the same dataset. As with the full model, two separate models were trained to predict outcomes at 24 months and over the entire follow-up period. Within the derivation set, LightAutoML performs model selection, hyperparameter tuning, and base-learner training using an internal K-fold cross-validation scheme: the training data are split into folds, and each base model is fitted on a subset of folds and used to generate out-of-fold (OOF) predictions for the held-out fold. These OOF predictions, which are unbiased estimates on the training data, are used both to optimize hyperparameters and to fit the higher-level blending/stacking layer that combines the individual algorithms. 

The held-out test set, defined by the patient-level split described above, was kept entirely separate from this internal cross-validation procedure and was used only for the final performance assessment.

\subsubsection{Construction of the \PreAFnine Linear Model and Tabular Risk-Score Derivation}

To further improve interpretability and facilitate the creation of a user-friendly risk score, we trained a simplified multivariate linear model using 13 features from our shortlist. No regularization was applied to avoid biasing the coefficient estimates, allowing for a more straightforward interpretation. 
After reviewing the resulting coefficients, we further reduced the feature set to the 9 most informative predictors for both time horizons: 24 months and the entire follow-up period.
The coefficients from these final linear models were then normalized by dividing by the smallest absolute coefficient and subsequently rounded to the nearest integers. This process resulted in two tabular risk scores suitable for bedside use, presented in Table~\ref{tab:scoring_24} and Table~\ref{tab:scoring_inf} for the 24-month and the entire follow-up targets, respectively.
This deliberate simplification yields a compact, paper-based scoring tool that can be applied at the bedside without internet access or computational resources, and whose limited set of integer-weighted predictors is amenable to mnemonic recall.

\subsubsection{Model Performance Assessment}

Model predictive value was assessed using the area under the receiver operating characteristic curve (ROC AUC); the area under the precision-recall curve (PRC-AUC); balanced accuracy of binary classification; individual recall, precision, and F1 score calculated with macro averaging. The pairwise comparison of the ROC AUC values was performed using DeLong's method, with z-statistics and P-values reported.

\subsubsection{Established Clinical Risk Scores as Baselines}

To evaluate the performance of the developed models, we compared them to several established clinical risk scores using ROC AUC as the evaluation metric. Specifically, we considered the CHARGE-AF \cite{alonso2013simple}, C2HEST \cite{li2019simple}, MHS \cite{aronson2018risk}, and HAVOC \cite{kwong2017clinical} risk scores. For this comparison, we manually implemented each scoring system according to their published definitions. Since some of these scores do not inherently provide a probability estimate, we normalized them by dividing each individual score by the maximum possible score defined by that scale. This allowed us to obtain a probability-like value for consistent ROC AUC computation across all methods.

\subsection{Model Interpretation}

To interpret the contribution of individual features to model predictions, we used the SHapley Additive exPlanations (SHAP) method \cite{lundberg2017unified}. SHAP is a model-agnostic technique based on cooperative game theory, which explains the output of machine learning models by computing the contribution of each feature to a particular prediction. In this framework, each feature is treated as a ``player'' in a game where the ``payout'' is the model’s prediction. SHAP values represent the average marginal contribution of a feature across all possible combinations (coalitions) of features. This allows us to quantify how much each feature increases or decreases the predicted risk for an individual. We used SHAP summary plots to visualize the distribution of feature impacts in the dataset. These plots provide insights into both the direction (positive or negative effect) and magnitude of each feature’s influence on the model’s predictions.

\subsection{Ethical Considerations}
The study was approved by the Independent Ethics Committee for Clinical Research of the E.I. Chazov National Medical Research Center of Cardiology, Ministry of Health of the Russian Federation (Moscow, Russia). The requirement for informed consent was waived owing to the retrospective design of the study and the use of deidentified patient data.

\subsection{Role of the Funding Source}
The funders of the study had no role in study design, data collection, data analysis, data interpretation, or the writing of the report.

\section{Results}

\subsection{Characteristics of the Study Group}
The study analyzed \textit{80,576} records encompassing \textit{45,000} unique patients. 17,562 patients met the inclusion and exclusion criteria. Of these patients, 1438 (8.19\%) received a new diagnosis of AF throughout the follow-up visits.
Figure \ref{fig:follow_up} in the Supplementary section presents the distribution of the times of the follow-up visits among unique patients. The follow-up duration ranges up to approximately 3,000 days, with a median of 426 days and a mean of 707 days.

The baseline characteristics are summarized in Table~\ref{tab:target_entire_stats} and Supplementary Table~\ref{tab:target_24_stats}. 
We observed a higher risk of incident AF in older individuals, women, those with a higher body mass index, systolic blood pressure, creatinine level, supraventricular beats number, left atrial volume, heart failure, valvular heart disease, left ventricular hypertrophy, 1-st AV block, COPD, diabetes, loop diuretic users, and unexpectedly, ARA 2 blockers (presumably as a heart failure surrogate). We observed a lower risk of new-onset AF in patients with previous percutaneous coronary intervention (PCI), in clopidogrel users, and, surprisingly, in patients with a previous myocardial infarction (MI). However, this association with MI was evident only for the entire follow-up period task (possibly due to more rigorous risk factor correction following such a diagnosis) and was not observed at the 24-month short-term period.

\begin{table}[htbp]
\centering
\small
\renewcommand{\arraystretch}{1.2}
\caption{Descriptive statistics for AF positive and negative groups for the entire follow-up period  \textit{}}
\label{tab:target_entire_stats}
\begin{tabularx}{\textwidth}{>{\raggedright\arraybackslash}p{2.5cm} X X X p{1.4cm}}
\hline
 & Overall (n=17562) & AF positive (n=1438) & AF negative (n=16124) & P-value \\
\hline
\textbf{Demographics} \\
\hline
Age, years & 62.0 [54.0, 70.0] & 67.0 [59.0, 75.0] & 62.0 [54.0, 69.0] & $<$0.001 \\
Female, n & 6396 (36.4) & 571 (39.7) & 5825 (36.1) & 0.007 \\
BMI, kg/m2 & 28.7 [25.7, 32.0] & 28.7 [25.7, 32.4] & 28.7 [25.7, 32.0] & 0.561 \\
\hline
\textbf{Comorbidities} \\
\hline
Hypertension, n & 14475 (82.4) & 1190 (82.8) & 13285 (82.4) & 0.745 \\
Systolic blood pressure, mm Hg & 130.0 [120.0, 140.0] & 130.0 [120.0, 145.0] & 130.0 [120.0, 140.0] & $<$0.001 \\
Diabetes, n & 3770 (21.5) & 359 (25.0) & 3411 (21.2) & $<$0.001 \\
Myocardial infarction, n & 8164 (46.5) & 594 (41.3) & 7570 (46.9) & $<$0.001 \\
Percutaneous coronary intervention, n & 8812 (50.2) & 565 (39.3) & 8247 (51.1) & $<$0.001 \\
Heart failure, n & 2884 (16.4) & 400 (27.8) & 2484 (15.4) & $<$0.001 \\
Valvular heart disease, n & 987 (5.6) & 156 (10.8) & 831 (5.2) & $<$0.001 \\
Left ventricle hypertrophy, n & 2935 (18.2) & 298 (20.7) & 2637 (16.4) & $<$0.001 \\
Left atrium volume, ml & 60.0 [50.0, 72.0] & 68.0 [57.0, 83.0] & 60.0 [50.0, 70.0] & $<$0.001 \\
First-degree AV block, n & 1893 (10.8) & 269 (18.7) & 1624 (10.1) & $<$0.001 \\
Supraventricular premature beats number per 24 hours, n & 38.0 [13.0, 143.0] & 130.0 [38.0, 681.5] & 34.0 [12.0, 121.0] & $<$0.001 \\
Sleep apnea, n & 331 (1.9) & 24 (1.7) & 307 (1.9) & 0.613 \\
COPD, n & 703 (4.0) & 83 (5.8) & 620 (3.8) & $<$0.001 \\
Hyperthyroidism, n & 48 (0.3) & 6 (0.4) & 42 (0.3) & 0.284 \\
Creatinine,  $\mu$mol/L & 79.0 [69.0, 93.7] & 82.0 [70.5, 98.0] & 79.0 [69.0, 93.0] & $<$0.001 \\
Potassium, mmol/L & 4.6 [4.2, 4.9] & 4.6 [4.3, 4.9] & 4.6 [4.2, 4.9] & 0.117 \\
\hline
\textbf{Medications} \\
\hline
ARBs, n & 3766 (21.4) & 370 (25.7) & 3396 (21.1) & $<$0.001 \\
Clopidogrel, n & 8399 (47.8) & 501 (34.8) & 7898 (49.0) & $<$0.001 \\
Loop diuretics, n & 2267 (12.9) & 319 (22.2) & 1948 (12.1) & $<$0.001 \\
\hline
\end{tabularx}
\end{table}

\subsection{Information Extraction Performance}

This section discusses the performance of the NLP pipeline for extracting features from the texts of discharge summaries. Note that here, we do not cover the training and testing of the AF diagnostic models.

\subsubsection{Rule-Based Extraction}

Across binary features, rule-based parser achieved a mean F1 score of 0.98 (range: 0.93--1.00). Three
features -- coronary artery bypass surgery, obstructive sleep apnoea syndrome,
and supraventricular tachycardia -- were extracted with a perfect F1 score of
1.00, while the remaining ones scored between 0.93 and 0.99, with heart failure
showing the lowest performance (F1 = 0.93). The continuous and
measurement-based variables, evaluated by extraction accuracy, scored between
0.89 and 1.00 (mean 0.97); age, sex, body mass index, and left atrial volume
were extracted with 100\% accuracy, whereas the supraventricular ectopic-beat
count was the most difficult (accuracy 0.89). The average performance of the
rule-based parser across all features was 0.97. We also note that the performance of extracting the target variable AF was also high, close to a 0.99 F1-score.

\subsubsection{ML-Based Named Entity Recognition}
 
For clinical features requiring contextual interpretation beyond keyword matching, transformer-based NER models were trained using a domain-adapted RuBioRoBERTa model. Three extraction tasks were addressed: drug name recognition, left ventricular hypertrophy (LVH) detection, and peripheral artery disease (PAD) mention extraction.
 
For drug name extraction, we compared three pretrained encoder-based language models on a held-out test set of 219 clinical documents containing 1{,}605 drug mention entities.
Domain adaptation yielded substantial improvements: the RuBioRoBERTa model \cite{yalunin2022rubioroberta} achieved the highest entity-level F1 of 0.95, compared with 0.93 for ruBERT-Sberbank and 0.91 for ruBERT-DeepPavlov. An additional entity linking step -- matching extracted mentions to a curated dictionary of cardiovascular medications via fuzzy string matching -- further improved F1 to 0.97 by filtering out spurious non-drug entities. Note that drug names extracted using a transformer-based model cover 28 features related to drug names from the full feature set.
 
For LVH detection from instrumental examination sections, the NER model achieved a
F1 score of 0.98 (178 test documents). For PAD mention extraction, evaluated on 404 test documents, the model reached a
F1 score of 0.92.

An average performance of the transformer-based extractors (with entity linking) is 0.95.

\subsubsection{Summary}

Performance for each feature is summarized in Supplementary Table~\ref{tab:nlpperf}.
Both rule-based and transformer-based extractors achieve individually high performance.
Taken together, the combined NLP pipeline achieves a 0.97 macro averaged performance score, indicating that the feature extraction performance is suitable for downstream clinical prediction modeling.

\subsection{Atrial Fibrillation Prediction Models}

\subsubsection{Performance of AF Prediction Models}

\begin{figure}[htbp]
    \centering
    \includegraphics[width=\textwidth]{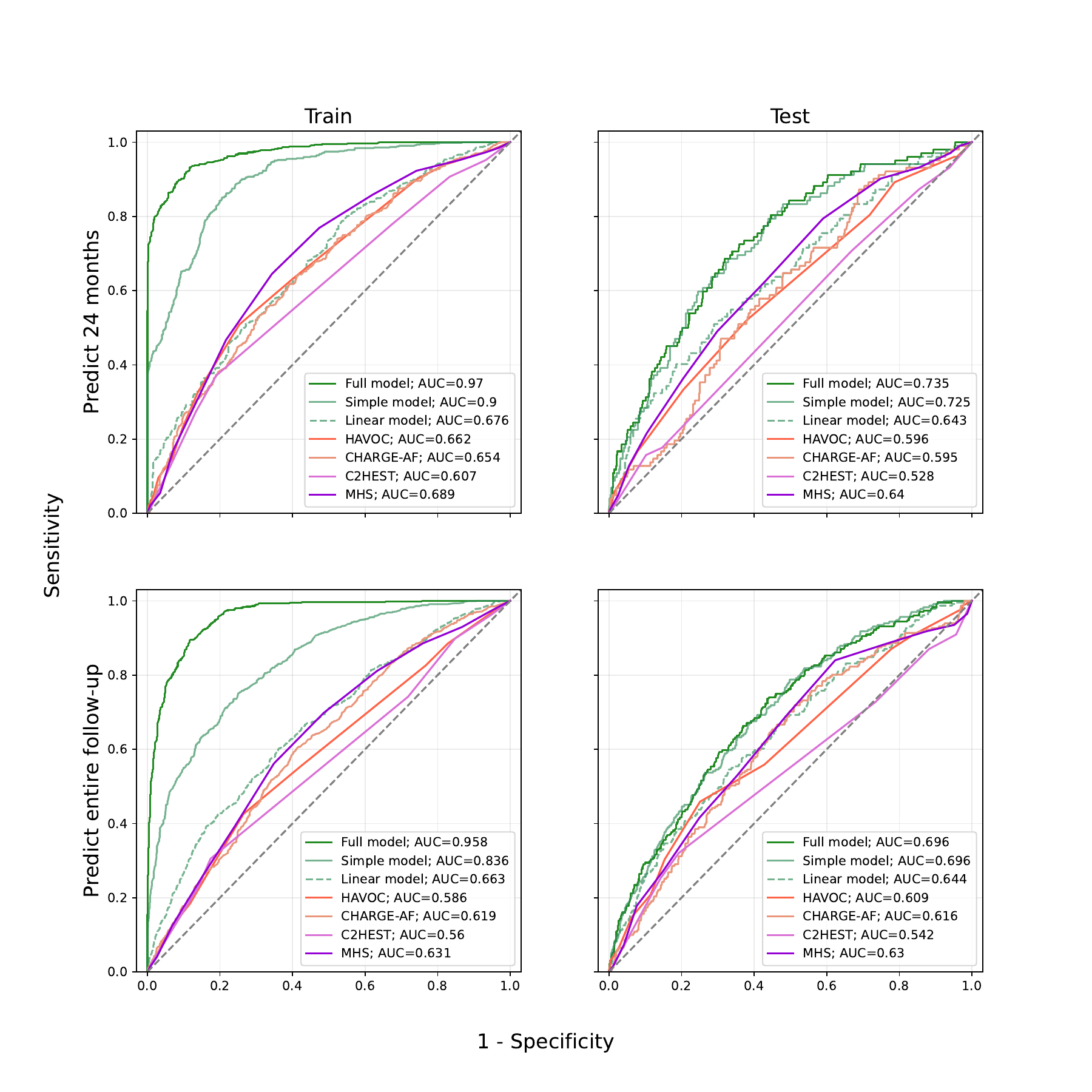}
    \caption{Receiver Operating Characteristic (ROC) curves comparing different models for predicting outcomes at different time points. The Area Under the Curve (AUC) values are reported for each model.}
    \label{fig:roc_comparison}
\end{figure}

The receiver operating characteristic curves for AF prediction models on both the training and test sets are presented in Figure \ref{fig:roc_comparison}. Extended evaluation metrics are provided in  Table \ref{tab:metrics_inf}
for the entire follow-up target and Table \ref{tab:metrics_24}
for the 24-month target.

\begin{table}[t]
\centering
\caption{24 months risk classification performance metrics.}
\label{tab:metrics_24}
\resizebox{\textwidth}{!}{%
\begin{tabular}{l *{12}{c}}
\toprule
{} & \multicolumn{2}{c}{Balanced Accuracy} & \multicolumn{2}{c}{F1 (macro averaging)} & \multicolumn{2}{c}{Precision (macro averaging)} & \multicolumn{2}{c}{Recall (macro averaging)} & \multicolumn{2}{c}{ROC AUC} & \multicolumn{2}{c}{PRC AUC} \\
\cmidrule(lr){2-3} \cmidrule(lr){4-5} \cmidrule(lr){6-7} \cmidrule(lr){8-9} \cmidrule(lr){10-11} \cmidrule(lr){12-13}
{} & Test & Train & Test & Train & Test & Train & Test & Train & Test & Train & Test & Train \\
\midrule
Full model   & 0.655 & 0.904 & 0.561 & 0.904 & 0.566 & 0.904 & 0.655 & 0.904 & 0.735 & 0.970 & 0.233 & 0.974 \\
Simple model & 0.674 & 0.821 & 0.543 & 0.821 & 0.565 & 0.822 & 0.674 & 0.821 & 0.725 & 0.900 & 0.220 & 0.903 \\
Linear model & 0.500 & 0.500 & 0.083 & 0.333 & 0.045 & 0.250 & 0.500 & 0.500 & 0.643 & 0.676 & 0.173 & 0.674 \\
HAVOC        & 0.500 & 0.500 & 0.476 & 0.333 & 0.455 & 0.250 & 0.500 & 0.500 & 0.596 & 0.662 & 0.136 & 0.650 \\
CHARGE-AF    & 0.500 & 0.500 & 0.476 & 0.333 & 0.455 & 0.250 & 0.500 & 0.500 & 0.595 & 0.654 & 0.137 & 0.629 \\
C2HEST       & 0.500 & 0.500 & 0.476 & 0.333 & 0.455 & 0.250 & 0.500 & 0.500 & 0.528 & 0.607 & 0.102 & 0.600 \\
MHS          & 0.500 & 0.500 & 0.476 & 0.333 & 0.455 & 0.250 & 0.500 & 0.500 & 0.640 & 0.689 & 0.147 & 0.650 \\
\bottomrule
\end{tabular}%
}
\end{table}

\begin{table}[t]
\centering
\caption{Entire follow up time risk classification performance metrics.}
\label{tab:metrics_inf}
\resizebox{\textwidth}{!}{%
\begin{tabular}{l *{12}{c}}
\toprule
{} & \multicolumn{2}{c}{Balanced Accuracy} & \multicolumn{2}{c}{F1 (macro averaging)} & \multicolumn{2}{c}{Precision (macro averaging)} & \multicolumn{2}{c}{Recall (macro averaging)} & \multicolumn{2}{c}{ROC AUC} & \multicolumn{2}{c}{PRC AUC} \\
\cmidrule(lr){2-3} \cmidrule(lr){4-5} \cmidrule(lr){6-7} \cmidrule(lr){8-9} \cmidrule(lr){10-11} \cmidrule(lr){12-13}
{} & Test & Train & Test & Train & Test & Train & Test & Train & Test & Train & Test & Train \\
\midrule
Full model   & 0.623 & 0.880 & 0.535 & 0.880 & 0.548 & 0.890 & 0.623 & 0.880 & 0.696 & 0.958 & 0.182 & 0.955 \\
Short model  & 0.630 & 0.738 & 0.511 & 0.737 & 0.544 & 0.742 & 0.630 & 0.738 & 0.696 & 0.836 & 0.178 & 0.841 \\
Linear model & 0.500 & 0.500 & 0.077 & 0.333 & 0.541 & 0.250 & 0.500 & 0.500 & 0.644 & 0.663 & 0.147 & 0.659 \\
HAVOC        & 0.500 & 0.500 & 0.478 & 0.333 & 0.459 & 0.250 & 0.500 & 0.500 & 0.609 & 0.586 & 0.138 & 0.579 \\
CHARGE-AF    & 0.499 & 0.500 & 0.478 & 0.333 & 0.459 & 0.250 & 0.499 & 0.500 & 0.616 & 0.619 & 0.115 & 0.596 \\
C2HEST       & 0.500 & 0.500 & 0.478 & 0.333 & 0.459 & 0.250 & 0.500 & 0.500 & 0.542 & 0.560 & 0.106 & 0.576 \\
MHS          & 0.500 & 0.500 & 0.478 & 0.333 & 0.459 & 0.250 & 0.500 & 0.500 & 0.630 & 0.631 & 0.122 & 0.598 \\
\bottomrule
\end{tabular}%
}
\end{table}

The full model achieved a ROC AUC of 0.735 for the 24-month period and 0.696 for the entire follow-up on the test set. The simple model yielded a slightly lower value of 0.725 for the 24-month target and a nearly identical ROC AUC of 0.696 for the entire follow-up. In all cases, our non-linear models exhibited superior discriminative performance relative to the reference risk scores: HAVOC, CHARGE-AF, C2HEST, and MHS.

As expected, the linear model used to derive the scoring system showed reduced performance, with ROC AUC values of 0.643 for the 24-month period and 0.644 for the entire follow-up. In all cases, our linear model was at least non‑inferior to all reference risk scores.

\subsubsection{A New \PreAF Scoring Method}

The predictors of AF that entered the simple nonlinear model were age, left atrial volume, the number of supraventricular ectopic beats per 24 hours, heart failure, prior PCI, serum creatinine and potassium levels, echocardiographic and/or ECG signs of left ventricular hypertrophy, systolic blood pressure at initial presentation, first‑degree atrioventricular block, prior myocardial infarction, use of clopidogrel, and use of loop diuretics. Given the number of significant factors in the simple nonlinear model, which we consider the most suitable for further validation and testing in real‑world clinical practice, we have named it \PreAF. 

\subsubsection{Feature Importance Analysis}

\begin{figure}[htbp]
    \centering
    \includegraphics[width=0.8\textwidth]{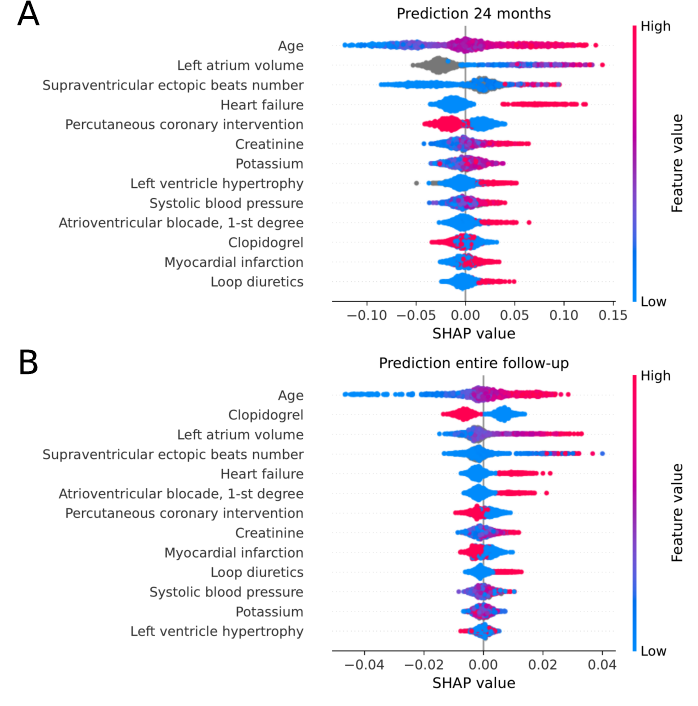}
    \caption{SHAP summary plots for feature importance in predicting risk at 24 months \textbf{(A)} and entire follow-up period \textbf{(B)}. The x-axis represents the SHAP values, indicating the impact of each feature on the model's prediction. Features are ranked by importance, with higher absolute SHAP values contributing more significantly. Colors indicate feature values, where red represents higher values and blue represents lower values.}
    \label{fig:feat_importance}
\end{figure}

Here, we present the results of applying an interpretability technique to the simple AF prediction models. Figure \ref{fig:feat_importance} displays the Shapley values for the input features of the simple model, ranked in descending order of importance. As expected, age is the most influential feature in predicting AF risk for both follow-up periods. The figure also illustrates how specific feature values are associated with increased or decreased AF risk. Notably, clopidogrel administration emerged as a protective factor and ranked as the second most important feature in the model for the entire follow-up period, although its contribution was substantially lower in the 24-month model. Elevated values of left atrial volume and supraventricular ectopic beats were associated with increased predicted AF risk, indicating that higher values of these features lead to a higher predicted AF risk.

\begin{figure}[htbp]
    \centering
    \includegraphics[width=0.8\textwidth]{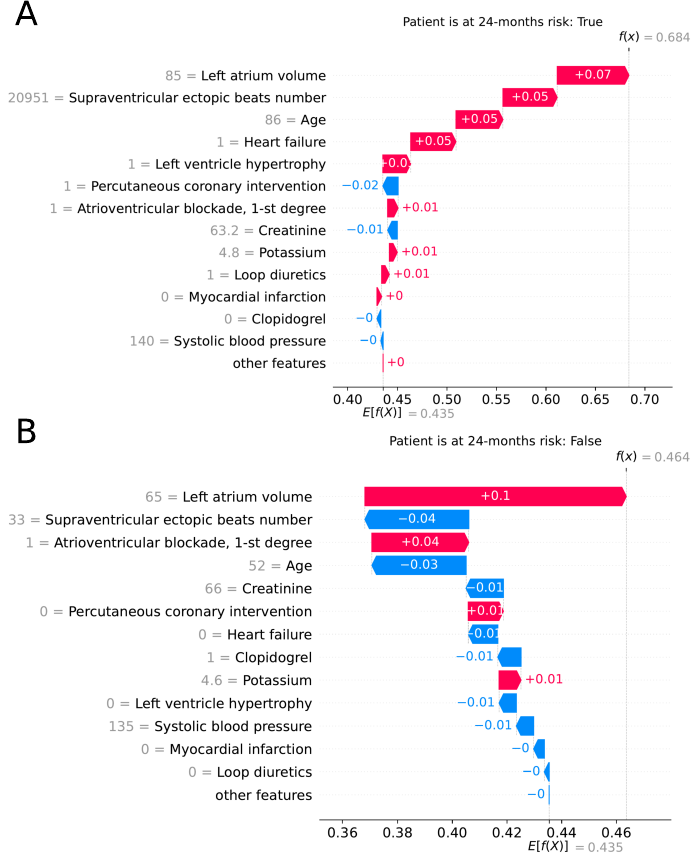}
    \caption{Shapley additive explanations illustrating feature contributions to the 24-month AF risk prediction for individuals. \textbf{(A)} A patient correctly classified as being at risk within 24 months (\textit{True}). The predicted risk is $f(x) = 0.684$, with key contributing features including left atrium volume, supraventricular ectopic beats number, age, and heart failure. \textbf{(B)} A patient correctly classified as not being at risk (\textit{False}), with a predicted risk of $f(x) = 0.464$. Left atrium volume acts as a risk-increasing factor, but age, supraventricular ectopic beats number, and other factors contribute negatively, keeping the overall risk below threshold.}
    \label{fig:feat_individuals}
\end{figure}

Since the simple model is non-linear, the contribution of individual features can vary across patients. To illustrate this, we visualized feature-level contributions for two patients with different predicted risks according to the 24-month follow-up model (Figure \ref{fig:feat_individuals}). The feature influence values were computed using the SHAP method. Features that increase AF probability are shown in red, while those that reduce it are shown in blue. Notably, left atrial volume contributes differently to the predictions for two patients of different ages. Furthermore, age itself acts as a risk-enhancing factor for one patient (Fig. \ref{fig:feat_individuals}A) and as a protective factor for another (Fig. \ref{fig:feat_individuals}B).

\textbf{}

\subsection{A Simplified \PreAFnine Scoring Method}

Based on the linear model of AF risk, a “classical” scale was developed that can be used without an electronic interface. We developed a scoring system based on individual risk factors (features) for both prediction targets. For the entire follow-up period, the corresponding scoring table (Table \ref{tab:scoring_inf}) includes 9 risk factors: Age, $\log_{10}$ of supraventricular ectopic beats number ($\log_{10}(\text{SVE})$), left atrial volume (LAVol), creatinine level, systolic blood pressure, heart failure, 1-st degree atrioventricular block, clopidogrel administration, and prior percutaneous coronary intervention. The relationship between the total assigned score and the absolute observed risk of atrial fibrillation is presented in Table \ref{tab:scoring_inf_incidence}.

\begin{table}[h]
\caption{Points assigned to entire follow up time AF risk factor categories (\PreAFnine score)}
\label{tab:scoring_inf}
\centering
\begin{tabularx}{\linewidth}{>{\raggedright\arraybackslash}p{7cm} X}
\rowcolor[HTML]{C0C0C0} 
Characteristics                        & Score \\ \hline
\rowcolor[HTML]{C0C0C0} 
Age, years                                    &       \\
\textless{}20                          & 0     \\
20-29                                  & 2     \\
29-38                                  & 5     \\
38-47                                  & 7     \\
47-55                                  & 10    \\
55-64                                  & 12    \\
64-73                                  & 15    \\
73-82                                  & 17    \\
\textgreater{}82                       & 20    \\
\rowcolor[HTML]{C0C0C0} 
log10(SVE)                             &       \\
\textless{}1.1                         & 4     \\
1.1-2.2                                & 8     \\
2.2-3.4                                & 12    \\
3.4-4.5                                & 17    \\
\rowcolor[HTML]{C0C0C0} 
LA volume, mL                              &       \\
\textless{}40                          & 2     \\
40-57                                  & 5     \\
57-76                                  & 7     \\
76-94                                  & 10    \\
94-112                                 & 12    \\
112-131                                & 15    \\
131-149                                & 17    \\
\textgreater{}149                      & 20    \\
\rowcolor[HTML]{C0C0C0} 
Creatinine, $\mu$mol/L                             &       \\
\textless{}96                          & 1     \\
96-148                                 & 3     \\
\textgreater{}148                      & 5     \\
\rowcolor[HTML]{C0C0C0} 
Systolic blood pressure, mm Hg                &       \\
80-160                                 & 1     \\
\textgreater{}160                      & 2     \\
\rowcolor[HTML]{C0C0C0} 
Heart failure                 & 4     \\
\rowcolor[HTML]{C0C0C0} 
1-st degree AV block & 4     \\
\rowcolor[HTML]{C0C0C0} 
Clopidogrel         & -5    \\
\rowcolor[HTML]{C0C0C0} 
PCI                                    & -2   
\end{tabularx}
\end{table}
\begin{table}
\centering
\caption{Predicted entire follow up time risk of AF by \PreAFnine score}
\label{tab:scoring_inf_incidence}
\begin{tabularx}{\linewidth}{>{\raggedright\arraybackslash}p{1.5cm} p{1.5cm} p{1.5cm} X}
\rowcolor[HTML]{C0C0C0} 
Risk score &  N. of subjects &  N. of AF incidences &     \% of subjects with AF \\
\hline
     0-19 &  10335 &  556 & 5.4 \\
    20-29 &   5562 &  567 & 10.2 \\
    30-39 &   1484 &  271 & 18.3 \\
      $\geq$40 &    158 &   40 & 25.3 \\
\end{tabularx}
\end{table}

Similarly, for the 24-month follow-up period, the scoring table (Table \ref{tab:scoring_24}) also includes 9 risk factors: Age, $\log_{10}(\text{SVE})$, LAVol, creatinine level, systolic blood pressure, heart failure, left ventricular hypertrophy, loop diuretic administration, and prior percutaneous coronary intervention. The corresponding association between total score and observed AF risk is shown in Table \ref{tab:scoring_24_incidence}. The differences in selected risk factors between the two scoring systems are due to the significance of their respective coefficients in the linear models used to derive the scores. For comparison, the observed AF incidence stratified by score category for the four established scales is shown in Supplementary Tables~\ref{tab:havoc_24_incidence}--\ref{tab:whs_24_incidence} (24-month target) and Supplementary Tables~\ref{tab:havoc_inf_incidence}--\ref{tab:whs_inf_incidence} (entire follow-up).

\begin{table}[]
\caption{Points assigned to 24-months AF risk factor categories (\PreAFnine score)}
\label{tab:scoring_24}
\centering
\begin{tabularx}{\linewidth}{>{\raggedright\arraybackslash}p{7cm} X}
\rowcolor[HTML]{C0C0C0} 
Characteristics          & Score \\ \hline
\rowcolor[HTML]{C0C0C0} 
Age, years                      &       \\
\textless{}20            & 0     \\
20-27                    & 2     \\
27-34                    & 4     \\
34-41                    & 6     \\
41-48                    & 9     \\
48-55                    & 11    \\
55-62                    & 13    \\
62-69                    & 15    \\
69-76                    & 17    \\
76-83                    & 20    \\
\textgreater{}83         & 22    \\
\rowcolor[HTML]{C0C0C0} 
log10(SVE)               &       \\
\textless{}0.7           & 3     \\
0.7-1.3                  & 6     \\
1.3-2.0                  & 9     \\
2.0-2.7                  & 12    \\
2.7-3.3                  & 16    \\
\textgreater{}3.3        & 19    \\
\rowcolor[HTML]{C0C0C0} 
LA volume, mL                 &       \\
\textless{}40            & 1     \\
40-60                    & 2     \\
60-80                    & 4     \\
80-120                   & 6     \\
\textgreater{}120        & 8     \\
\rowcolor[HTML]{C0C0C0} 
Creatinine, $\mu$mol/L            &       \\
\textless{}82            & 1     \\
82-120                   & 3     \\
120-160                  & 4     \\
\textgreater{}160        & 6     \\
\rowcolor[HTML]{C0C0C0} 
Systolic blood pressure, mm Hg  &       \\
90-135                   & 1     \\
135-170                  & 2     \\
\textgreater{}170        & 4     \\
\rowcolor[HTML]{C0C0C0} 
Heart failure   & 7     \\
\rowcolor[HTML]{C0C0C0} 
LVH                      & 4     \\
\rowcolor[HTML]{C0C0C0} 
Loop diuretics & 2     \\
\rowcolor[HTML]{C0C0C0} 
PCI                      & -3
\end{tabularx}
\end{table}
\begin{table}
\centering
\caption{Predicted 24-months risk of AF by \PreAFnine score}
\label{tab:scoring_24_incidence}
\begin{tabularx}{\linewidth}{>{\raggedright\arraybackslash}p{1.5cm} p{1.5cm} p{1.5cm} X}
\rowcolor[HTML]{C0C0C0} 
Risk score &  N. of subjects &  N. of AF incidences &     \% of subjects  with AF\\
\hline
    0-24 &   4619 &  337 & 7.3 \\
   25-34 &   1644 &  223 & 13.6 \\
   35-44 &    479 &  142 & 29.6 \\
   45-49 &     54 &   18 & 33.3 \\
    $\geq$ 50 &     11 &    4 & 36.3 \\
\end{tabularx}
\end{table}

\section{Discussion}

\subsection{Advantages of the Developed Models}
 
In this study, we developed and internally validated machine learning models for predicting atrial fibrillation in patients with established cardiovascular disease, using 73 features extracted automatically from electronic health records using a custom NLP pipeline. The non-linear models achieved test ROC AUCs of 0.73--0.74 for 24 months and  0.70 for entire follow-up, consistently outperforming four established clinical risk scores evaluated on the same cohort. Notably, the 13-feature simple model performed nearly identically to the full model (ROC AUC difference $\leq$0.01), indicating that discriminative information is concentrated in a small set of clinically interpretable variables.

As references, we selected three well-validated risk scores (CHARGE-AF\cite{alonso2013simple}, C$_2$HEST\cite{li2019simple}, MHS\cite{aronson2018risk}) that have been endorsed for clinical use by professional societies \cite{gelder2024guidelines}. Another score chosen for comparison (HAVOC \cite{kwong2017clinical}) was developed for patients with established vascular pathology (cryptogenic stroke or transient ischemic attack), which, to some extent, rendered the characteristics of such patients similar to those of our cohort. The limited discriminative ability of CHARGE-AF, C$_2$HEST, MHS, and HAVOC in our population (ROC AUC 0.53--0.64) likely reflects the high prevalence of their constituent risk factors among hospitalized cardiovascular patients, leaving little residual stratification power. Our models address this by incorporating features that capture atrial substrate remodeling and treatment effects.  

Given we are predicting incident AF in a high‑risk population, the absolute values of the ROC AUC metric are modest; and these values comparable to the performance ranges reported in similar studies~\cite{pipilas2023use, alonso2013simple}.
 
Our results suggest that the dynamic nature of AF risk typically renders accurate long‑term risk estimation challenging, whereas a 24‑month risk assessment is more precise. This is typical for predicting risk in multifactorial cardiovascular diseases \cite{urbut2024msgene}. For some diseases (e.g., CAD), this is a drawback, but for AF, a shorter prediction horizon is more convenient in practice, as it restricts the patient focus group for active screening to the near future. To our knowledge, there are no other scores that have been specifically created or adapted for medium‑term AF prediction.

Moreover, several other features corroborate the potential clinical applicability of our models. The models rely exclusively on routinely collected hospital data without additional diagnostic procedures, enabling EHR-integrated screening. The derived tabular risk scores offer a simplified bedside tool, with a 24-month observed AF incidence ranging from approximately 7\% in the lowest to over 36\% in the highest risk category. SHAP-based explanations also help personalize risk communication, since showing patients their own risk -- especially with visuals -- is known to improve adherence to prevention \cite{naderian2025effect, naslund2019visualization}.

\subsection{AF Predictors}

The predictors identified by SHAP analysis are consistent with established pathophysiology. Age and left atrial volume -- the two dominant features -- reflect cumulative atrial fibrosis  and structural remodeling, respectively \cite{lin2021impaired} Supraventricular ectopy, first-degree atrioventricular block, and heart failure all reflect electrical instability or hemodynamic conditions favoring arrhythmogenesis. 

In the case of clopidogrel, two potential explanations can be proposed. First, the drug may exert a direct protective effect by suppressing hypercoagulation and through possible pleiotropic mechanisms \cite{kuszynski2022pleiotropic}. Experimental studies have previously shown that hypercoagulation \textit{per se} increases the risk of AF \cite{spronk2017hypercoagulability}. Second, clopidogrel administration typically followed percutaneous coronary intervention (PCI) and/or acute coronary syndrome (ACS); such patients generally receive closer follow‑up and demonstrate better treatment adherence, which secondarily leads to improved risk factor control.
The same mechanism -- more intensive patient management -- may also account for the apparent ``pseudo‑protective'' effect of myocardial infarction observed during long‑term follow‑up in our study.
With regard to myocardial infarction (MI), generally recognized as a predictor of AF, several other observational studies have failed to confirm its association with AF \cite{macfarlane2011incidence}. It is likely that the more significant predictor of AF is not MI \textit{per se}, but rather its severity in terms of subsequent hemodynamic impact. After adjustment for heart failure, diastolic dysfunction, left ventricular ejection fraction, and other relevant factors, the independent predictive value of MI was neutralized. Moreover, therapies for MI, such as early revascularization, can sometimes mitigate the risk of AF, which may reduce the prognostic significance of MI.

 A possible explanation for the protective effect of PCI is the relief of myocardial ischemia -- not only atrial but also ventricular, which influences left ventricular function and left atrial preload. Unlike coronary artery bypass grafting, PCI does not trigger AF in the acute postoperative period \cite{kosmidou2018new}. 

Differences in predictor rankings between time horizons suggest that markers of acute hemodynamic stress (LVH, loop diuretics) are more relevant for near-term prediction, while chronic treatment patterns and conduction disease carry greater weight over longer follow-up.

\subsection{Methodological Contribution}

A distinctive methodological contribution is the end-to-end NLP pipeline transforming unstructured Russian-language EHRs into analysis-ready features. The rule-based component achieved an average performance score of 0.98 across binary clinical features, while the transformer-based NER models attained an average F1 of 0.95, demonstrating that high-quality automated extraction from non-English clinical text is feasible without manual chart review. The use of active learning and domain-adapted pretraining (RuBioRoBERTa) proved essential for achieving this performance with limited expert annotation.

\subsection{Web-Based Prediction Tool}

The final models were integrated into an easy-to-use, interpretable web-based prediction tool designed to support clinical decision-making\footnote{Available at \url{https://cardio.airi.net/}.}. This online tool provides not only the estimated risk of AF development but also visually represents the individual contribution of risk factors to the prognosis for each patient, such as those shown in Fig. \ref{fig:feat_individuals}.

\subsection{Limitations}

This single-center retrospective study of hospitalized cardiovascular patients limits generalizability to primary care or outpatient settings. The target variable, based on documented AF diagnoses during subsequent visits, likely underestimates the true incidence by missing paroxysmal or subclinical episodes. The NLP pipeline, while performant, was evaluated on relatively small test sets and would require adaptation for other languages. The study does not incorporate ECG waveform data or cardiac biomarkers (e.g., NT-proBNP), which could improve performance. The gap between training and test ROC AUC values indicates overfitting that warrants attention through larger cohorts or stronger regularization. External validation and prospective evaluation are needed before clinical deployment.

\subsection{Conclusion}

This study demonstrates that machine-learning models applied to routinely available hospital data can effectively identify patients at heightened risk of developing atrial fibrillation. The greatest predictive value is achieved by a nonlinear model assessing the risk of AF over a 24‑month period. By leveraging both structured and unstructured information from electronic health records, the models capture clinically relevant predictors that align with established mechanisms of atrial fibrillation. Their superior performance compared with existing clinical scores highlights the potential for these tools to support more precise risk stratification and to guide timely diagnostic evaluation and screening strategies.

\section*{Conflict of Interest Statement}
The authors declare no conflict of interest.

\section*{Author Contributions}
O.S., E.P., and D.K. performed expert curation and annotation of medical data; D.K., D.L., N.K., M.K., and A.S. trained machine learning models; I.B., A.Z., K.G., E.I., I.S., and D.D. assisted with information retrieval tasks; O.S., D.K., A.S., and D.D. performed study design and analyzed the results; O.S., D.K., and A.S. prepared the initial draft of the manuscript; all authors participated in manuscript editing and proofreading. A.S. led the natural language processing, O.S., E.P., and D.D. formulated the hypothesis and acquired funding for the project.

\section*{Funding}
The project was supported by Russian Foundation for Basic Research (grant \#19-29-01240 ``Development of methods based on multimodal medical data analysis for predicting atrial fibrillation and cardioembolic ischemic stroke'').

\section*{Data Availability Statement}
The data is available to other researchers upon reasonable request sent to the corresponding authors.

\bibliographystyle{plainnat}
\bibliography{ref}

\clearpage
\appendix
\section*{Supplementary Material}
\setcounter{figure}{0}
\setcounter{table}{0}
\renewcommand{\thefigure}{S\arabic{figure}}
\renewcommand{\thetable}{S\arabic{table}}
\begin{figure}[h]
    \centering
    \includegraphics[width=0.9\linewidth]{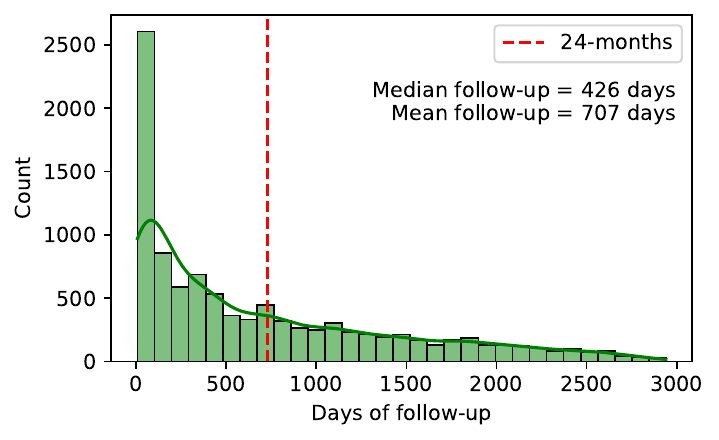}
    \caption{Distribution of follow-up time among unique patients. The follow-up duration ranges up to approximately 3000 days, with a median of 426 days and a mean of 707 days. A dashed line indicates the 24-month (730-day) mark.}
    \label{fig:follow_up}
\end{figure}

\clearpage
\newpage
\begin{table}[h]
\centering
\caption{Predicted 24-months risk of atrial fibrillation by HAVOC score}
\label{tab:havoc_24_incidence}
\begin{tabularx}{\linewidth}{>{\raggedright\arraybackslash}p{1.5cm} p{1.5cm} p{1.5cm} X}
\rowcolor[HTML]{C0C0C0} 
Risk score &  N. of subjects &  N. of AF incidences &     \% of subjects with AF \\
\hline
  0-1 &    320 &   16 & 5.00 \\
  2-3 &   1465 &   66 & 4.50 \\
  4-6 &   4061 &  418 & 10.3 \\
  7-8 &    596 &  124 & 20.8 \\
 9-11 &    341 &   88 & 25.8 \\
  $>$11 &     34 &   13 & 38.2 \\
\end{tabularx}
\end{table}

\begin{table}
\centering
\caption{Predicted 24-months risk of atrial fibrillation by CHARGE-AF score}
\label{tab:charge_24_incidence}
\begin{tabularx}{\linewidth}{>{\raggedright\arraybackslash}p{1.5cm} p{1.5cm} p{1.5cm} X}
\rowcolor[HTML]{C0C0C0} 
Risk score &  N. of subjects &  N. of incidences &     \% of subjects \\
\hline
     0-5\% &   6145 &  579 & 9.4 \\
    5-10\% &    383 &   75 & 19.6 \\
   10-15\% &    161 &   38 & 23.6 \\
   15-20\% &     65 &   18 & 27.7 \\
     $>$20\% &     63 &   15 & 23.8 \\
\end{tabularx}
\end{table}

\begin{table}
\centering
\caption{Predicted 24-months risk of atrial fibrillation by C$_2$HEST score}
\label{tab:c2hest_24_incidence}
\begin{tabularx}{\linewidth}{>{\raggedright\arraybackslash}p{1.5cm} p{1.5cm} p{1.5cm} X}
\rowcolor[HTML]{C0C0C0} 
Risk score &  N. of subjects &  N. of incidences &     \% of subjects \\
\hline
     0 &    578 &   34 & 5.98 \\
     1 &   1171 &   96 & 8.2 \\
     2 &   3401 &  309 & 9.1 \\
     3 &    341 &   63 & 18.5 \\
     4 &    775 &  169 & 21.8 \\
     5 &     56 &   12 & 21.4 \\
    $\geq$6 &     28 &    5 & 17.8 \\
\end{tabularx}
\end{table}

\begin{table}
\centering
\caption{Predicted 24-months risk of atrial fibrillation by MHS score}
\label{tab:whs_24_incidence}
\begin{tabularx}{\linewidth}{>{\raggedright\arraybackslash}p{1.5cm} p{1.5cm} p{1.5cm} X}
\rowcolor[HTML]{C0C0C0} 
Risk score &  N. of subjects &  N. of incidences &     \% of subjects \\
\hline
 -1 &    277 &    9 & 3.2 \\
  0 &    514 &   17 & 3.3 \\
1-2 &   1653 &   72 & 4.3 \\
3-5 &   1790 &  164 & 9.2 \\
6-7 &   1400 &  241 & 17.2 \\
8-9 &    764 &  172 & 22.5 \\
$\geq$10 &    193 &   39 & 20.2 \\
\end{tabularx}
\end{table}

\begin{table}
\centering
\caption{Predicted entire follow-up time risk of AF by HAVOC score}
\label{tab:havoc_inf_incidence}
\begin{tabularx}{\linewidth}{>{\raggedright\arraybackslash}p{1.5cm} p{1.5cm} p{1.5cm} X}
\rowcolor[HTML]{C0C0C0} 
Risk score &  N. of subjects &  N. of AF incidences &     \% of subjects with AF \\
\hline
  0-1 &    613 &   31 & 5.1 \\
  2-3 &   3225 &  209 & 6.5 \\
  4-6 &  10881 &  812 & 7.4 \\
  7-8 &   1687 &  211 & 12.5 \\
 9-11 &   1038 &  147 & 14.2 \\
  $>$11 &    118 &   28 & 23.7 \\
\end{tabularx}
\end{table}

\begin{table}
\centering
\caption{Predicted entire follow-up time risk of atrial fibrillation by CHARGE-AF score}
\label{tab:charge_inf_incidence}
\begin{tabularx}{\linewidth}{>{\raggedright\arraybackslash}p{1.5cm} p{1.5cm} p{1.5cm} X}
\rowcolor[HTML]{C0C0C0} 
Risk score &  N. of subjects &  N. of incidences &     \% of subjects \\
\hline
     0-5\% &  15861 & 1174 & 7.4 \\
    5-10\% &    991 &  139 & 14.02 \\
   10-15\% &    422 &   70 & 16.6 \\
   15-20\% &    140 &   34 & 24.3 \\
     $>$20\% &    148 &   21 & 14.2 \\
\end{tabularx}
\end{table}

\begin{table}
\centering
\caption{Predicted entire follow-up time risk of atrial fibrillation by C$_2$HEST score}
\label{tab:c2hest_inf_incidence}
\begin{tabularx}{\linewidth}{>{\raggedright\arraybackslash}p{1.5cm} p{1.5cm} p{1.5cm} X}
\rowcolor[HTML]{C0C0C0} 
Risk score &  N. of subjects &  N. of incidences &     \% of subjects \\
\hline
     0 &   1603 &   84 & 5.2 \\
     1 &   2236 &  222 & 9.9 \\
     2 &   9599 &  621 & 6.5 \\
     3 &    800 &  107 & 13.4 \\
     4 &   2136 &  305 & 14.3 \\
     5 &    148 &   22 & 14.9 \\
    $\geq$6 &     81 &    8 & 9.9 \\
\end{tabularx}
\end{table}

\begin{table}
\centering
\caption{Predicted entire follow up time risk of atrial fibrillation by MHS score}
\label{tab:whs_inf_incidence}
\begin{tabularx}{\linewidth}{>{\raggedright\arraybackslash}p{1.5cm} p{1.5cm} p{1.5cm} X}
\rowcolor[HTML]{C0C0C0} 
Risk score &  N. of subjects &  N. of incidences &     \% of subjects \\
\hline
 -1 &    575 &   28 & 4.9 \\
  0 &   1180 &   51 & 4.3 \\
1-2 &   4150 &  161 & 3.9 \\
3-5 &   4807 &  373 & 7.8 \\
6-7 &   3678 &  420 & 11.4 \\
8-9 &   2194 &  314 & 14.3 \\
$\geq$10 &    532 &   66 & 12.4 \\
\end{tabularx}
\end{table}

\begin{table*}[h]
\centering
\small
\renewcommand{\arraystretch}{1.0}
\caption{Descriptive statistics for AF positive and negative groups for the 24-months follow-up}
\label{tab:target_24_stats}
\begin{tabularx}{\textwidth}{>{\raggedright\arraybackslash}p{2.5cm} X X X p{1.4cm}}
\hline
 & Overall (n=6817) & AF positive (n=725) & AF negative (n=6092) & P-value \\
\midrule
Age, years & 62.0 [53.0, 71.0] & 68.0 [59.0, 76.0] & 61.0 [53.0, 70.0] & $<$0.001 \\
Female, n & 2725 (40.0) & 285 (39.3) & 2440 (40.1) & 0.718 \\
BMI, kg/m2 & 28.7 [25.8, 32.1] & 28.7 [25.7, 32.4] & 28.7 [25.8, 32.0] & 0.982 \\
Hypertension, n & 5552 (81.4) & 595 (82.1) & 4957 (81.4) & 0.686 \\
Systolic blood pressure, mm Hg & 130.0 [120.0, 140.0] & 130.0 [120.0, 145.0] & 130.0 [120.0, 140.0] & 0.017 \\
Diabetes, n & 1461 (21.4) & 185 (25.5) & 1276 (20.9) & 0.005 \\
Myocardial infarction, n & 2854 (41.9) & 303 (41.8) & 2551 (41.9) & 1.0 \\
Percutaneous coronary intervention, n & 3241 (47.5) & 288 (39.7) & 2953 (48.5) & $<$0.001 \\
Heart failure, n & 1039 (15.2) & 240 (33.1) & 799 (13.1) & $<$0.001 \\
Valvular heart disease, n & 389 (5.7) & 86 (11.9) & 303 (5.0) & $<$0.001 \\
Left ventricle hypertrophy, n & 1043 (16.5) & 168 (23.2) & 875 (14.4) & $<$0.001 \\
Left atrium volume, mL & 60.0 [52.0, 72.0] & 68.0 [55.0, 85.0] & 60.0 [51.0, 70.0] & $<$0.001 \\
Atrioventricular block, 1-st degree, n & 722 (10.6) & 137 (18.9) & 585 (9.6) & $<$0.001 \\
Supraventricular premature beats (number per 24 hours), n & 38.0 [13.0, 146.0] & 188.0 [49.0, 887.0] & 33.0 [12.0, 115.0] & $<$0.001 \\
Sleep apnea, n & 137 (2.0) & 14 (1.9) & 123 (2.0) & 1.0 \\
Chronic obstructive pulmonary disease, n & 273 (4.0) & 44 (6.1) & 229 (3.8) & 0.004 \\
Hyperthyroidism, n & 15 (0.2) & 4 (0.6) & 11 (0.2) & 0.067 \\
Creatinine, $\mu$mol/L & 79.0 [68.7, 93.0] & 84.0 [71.0, 100.0] & 78.5 [68.5, 93.0] & $<$0.001 \\
Potassium, mmol/L & 4.6 [4.2, 4.9] & 4.6 [4.3, 4.9] & 4.5 [4.2, 4.9] & 0.003 
\\
Angiotensin II receptor antagonists, n & 1465 (21.5) & 205 (28.3) & 1260 (20.7) & $<$0.001 \\
Clopidogrel, n & 2771 (40.6) & 250 (34.5) & 2521 (41.4) & $<$0.001 \\
Loop diuretics, n & 870 (12.8) & 195 (26.9) & 675 (11.1) & $<$0.001 \\
\bottomrule
\end{tabularx}
\end{table*}

\begin{table}[ht]
\centering
\caption{Performance of the NLP pipeline for individual feature extraction. ``All drugs'' covers 28 features from the feature set corresponding to drug names (active substances).}
\label{tab:nlpperf}
\begin{tabular}{llcc}
\toprule
Feature & Method & Metric & Score \\
\midrule
Age & Rules & Accuracy & 1.00 \\
All drugs & ML & F1 & 0.97 \\
Atrioventricular blockade, 1-st degree & Rules & F1 & 0.95 \\
BMI & Rules & Accuracy & 1.00 \\
Cholesterol & Rules & Accuracy & 0.96 \\
Chronic obstructive pulmonary disease & Rules & F1 & 0.97 \\
Coronary artery bypass surgery & Rules & F1 & 1.00 \\
Coronary artery disease & Rules & F1 & 0.99 \\
Creatinine & Rules & Accuracy & 0.96 \\
Diabetes & Rules & F1 & 0.99 \\
Diastolic blood pressure & Rules & Accuracy & 0.94 \\
Diastolic dysfunction of left ventricle & Rules & F1 & 0.98 \\
Female & Rules & Accuracy & 1.00 \\
Glucose & Rules & Accuracy & 0.96 \\
Heart failure & Rules & F1 & 0.93 \\
Heart rate & Rules & Accuracy & 0.96 \\
Height & Rules & Accuracy & 0.99 \\
Hemoglobin & Rules & Accuracy & 0.95 \\
Hypertension & Rules & F1 & 0.99 \\
Left atrium volume & Rules & Accuracy & 1.00 \\
Left ventricle hypertrophy & ML & F1 & 0.98 \\
Left ventricular ejection fraction & Rules & Accuracy & 0.99 \\
Left ventricular myocardial mass & Rules & Accuracy & 0.98 \\
Low density lipoproteins & Rules & Accuracy & 0.97 \\
Mean pulmonary pressure & Rules & Accuracy & 0.94 \\
Myocardial infarction & Rules & F1 & 0.99 \\
Obstructive sleep apnoea syndrome & Rules & F1 & 1.00 \\
Percutaneous coronary intervention & Rules & F1 & 0.97 \\
Peripheral artery disease & ML & F1 & 0.92 \\
Potassium & Rules & Accuracy & 0.97 \\
Previous stroke & Rules & F1 & 0.97 \\
Pulmonary embolism & Rules & F1 & 0.99 \\
Pulmonary hypertension & ML & F1 & 0.94 \\
Supraventricular ectopic beats number & Rules & Accuracy & 0.89 \\
Supraventricular tachycardia & Rules & F1 & 1.00 \\
Systemic thromboembolism & Rules & Accuracy & 0.97 \\
Systolic blood pressure & Rules & Accuracy & 0.96 \\
Thyroidal dysfunction & Rules & F1 & 0.98 \\
Transient ischemic attack & Rules & F1 & 0.94 \\
Triglycerides & Rules & Accuracy & 0.95 \\
Valvular disease & Rules & F1 & 0.95 \\
Weight & Rules & Accuracy & 0.96 \\
White blood cells count & Rules & Accuracy & 0.95 \\
\bottomrule
\end{tabular}

\end{table}

\end{document}